# Towards The Development of a Bishnupriya Manipuri Corpus

Nayan Jyoti Kalita, Navanath Saharia and Smriti Kumar Sinha
Department of Computer Science & Engineering
Tezpur University, Napaam
India – 784028
nayan.jk.123@gmail.com
{nava_tu,smriti}@tezu.ernet.in

**Abstract-** For any deep computational processing of language we need evidences, and one such set of evidences is corpus. This paper describes the development of a text-based corpus for the Bishnupriya Manipuri language. A Corpus is considered as a building block for any language processing tasks. Due to the lack of awareness like other Indian languages, it is also studied less frequently. As a result the language still lacks a good corpus and basic language processing tools. As per our knowledge this is the first effort to develop a corpus for Bishnupriya Manipuri language.

## I. INTRODUCTION

A Corpus is the source of data for the research of natural language processing applications. A corpus data may contain text in a single language (monolingual corpus) or in multiple languages (multilingual corpus). Newspapers and books are the most used resources for the development of a corpus. Now-a-days, the free resources on Internet are widely used as a source. A corpus should be a balanced one to cover all the properties of the different kinds of text it contains and should be a good representative for the language.

This paper reports the development of a corpus on the Bishnupriya Manipuri language and the result of the analysis of the data from the corpus. Section II gives an outline on the Bishnupriya Manipuri language. Section III describes the corpus development. Then the corpus is analysed in Section IV. The word analysis of Bishnupriya Manipuri language that we found from our corpus is also given in this section. Section V gives the analysis on word frequency. Section VI concludes the study followed by the references.

## II. THE BISHNUPRIYA MANIPURI LANGUAGE

The Manipuris are divided into two groups, namely, the Meiteis and the Bishnupriyas. The Bishnupriyas are of Aryo-Mongolo-Dravidian origin and the Meiteis are of Mongolian origin. The Bishnupriyas are generally dark in complexion. On the other hand, the Meiteis are generally yellowish in complexion. The Meiteis call their language 'Meitei' or 'Manipuri' which is the state-language of Manipur. Formerly, the Bishnupriyas used to call their language simply 'Manipuri' but, now, they call it 'Bishnupriya Manipuri' to distinguish it from Meitei. The Bishnupriya Manipuri language is of Indo-Aryan origin and is a kin to Oriya, Assamese and Bengali. On the other hand, the Meitei language is of the Kuki-Chin branch of the Tibeto-Burman group of languages. Though both the languages have differences in many factors, these two sections of people have a common stock of culture; their kirtana, dances, music, dress etc - all are of the same type. There is no bar to matrimonial relations between these sections of people, but in practice, it is of rare occurrence.

The Meitei language, though grouped under the Tibeto-Burman group of languages, has absorbed a number of words from the neighbouring group of Indo-Aryan languages. The Bishnupriya Manipuri language, on the other hand, though of Indo-Aryan origin, has incorporated numerous words (about 4,000) from Meitei. [1]

The Bishnupriya Manipuri language was originally confined only to the surroundings of the lake Loktak in Manipur. The principal localities where this language was spoken are now known as Khangabok, Heirok, Mayang-Yumphan, Bishnupur, Khunau, Ningthaukhong, Thamnapokpi and other places. The people of these places are known as Bishnupriyas even now, and are similar to the Bishnupriyas living outside Manipur in respect of their appearance and complexion. They, of course, neither speak nor understand Bishnupriya; they all speak Meitei. Formerly, Bishnupriya Manipuri speakers were very numerous in the localities mentioned, a major portion of which was included in the Khumal kingdom. But, when a great majority of these people fled from Manipur, during different reasons, it was very difficult for the few that remained there to retain their language in the face of the impact of Meitei spoken by the majority. They gradually began to forget their languages and assimilated with the speakers of the dominant language- The Meitei. There are two dialects in the Bishnupriya Manipuri language, namely, the Madai Gang dialect or the dialect of the village of the queen and the Rajar Gang dialect or the dialect of the village of the king. The Madai Gang dialect is also known as Leimanai and the Rajar Gang dialect as Ningthaunai. The term Leimanai is derived from Leima (queen)+(ma) nai (attendant), meaning the attendants of the queen, and the word Ningthaunai, from ningthau (king)+(ma) nai (attendant) meaning the attendants of the king. Unlike the dialects of other tribes, these dialects of Bishnupriya are not confined to distinct geographical areas; they rather exist side by side in the same localities [2].

The Bishnupriya Manipuri language is practically dead in its place of origin. However, the language is retained by its speakers in diaspora mostly in Assam, Tripura and Bangladesh. The language is still enlisted as an endangered language by UNESCO [3]. Interestingly, a thorough linguistic study of this little known language was done by Dr. K. P. Sinha. The study done was mainly on morphological analysis. An etymological dictionary is also published [4]. But no study on the computational linguistic of this language is done. To the best of our knowledge, the present work based on the available publications, is the first of its kind in this direction.

### III. CORPUS DEVELOPMENT

A corpus (plural *corpora*) or text corpus is a large collection of texts. They are used to do statistical analysis on linguistic rules on a specific universe.

A corpus may contain texts in a single language (*monolingual corpus*) or text data in multiple languages (*multilingual corpus*). Multilingual corpora that have been specially formatted for side-by-side comparison are called *aligned parallel corpora.*

In order to make the corpora more useful for doing linguistic research, they are often subjected to a process known as annotation. An example of annotating a corpus is *POS-tagging*, in which information about each word as a part of speech (verb, noun, adjective, etc.) is added to the corpus in the form of *tags*. When the language of the corpus is not a very known or popular language, the corpus is annotated bilingual.

Some corpora are further analysed in structural levels. In particular, a number of smaller corpora may be fully parsed. Such corpora are called Parsed Corpora or Tree banks. These corpora are completely and consistently annotated, and usually smaller, containing 1 to 3 million words. Corpora can be further analysed, including annotations for morphology, pragmatics etc.

Corpora are considered as the main knowledge base for any language processing task. Frequency list of words from the corpora are useful in linguistic works. They are also useful in language teaching. The analysis and processing of various types of corpora are widely done in speech recognition and machine translation. Corpora are very much helpful in learning and writing a foreign language because corpora allow the non-native language users to grasp the manner of sentence formation in the foreign language.

**Feature Analysis:**

While developing a corpus, we should consider some features. First, we should consider mood of the text data, whether the data collected originates in speech or writing. Second is the data samples collected for the corpus. Entire document or transcriptions of speech events might be considered for the corpus, but the samples should be in proper size so that the system can process the data. The third feature is that data samples collected for the corpus should be good enough for the representation of the language. The last one is that the corpus should be balanced. It should cover different kinds of texts with all properties.

It has already been realized that free resources on Internet such as blogs, Wikipedia, web pages etc. could be used for developing a large corpus. Such materials could create a problem when we collect them from different sources.

Free resources on Internet such as blogs, webpages, Wikipedia etc. become the source for the development of a corpus. But this method cannot help much for the Bishnupriya Manipuri language as very small number of people uses this language and very small numbers of resources are available on the Internet. Moreover, most of the resources present in the Internet are in graphic format, which cannot be used in this language processing work.

We have collected many texts from the Bishnupriya Manipuri version of Wikipedia [5]. Though it is a very small one, but still it could be used as a corpus. Further we have collected large number of text written in Smriti Font, which is an ASCII Font. For this reason, we have built an ASCII-to-UNICODE Converter, which converts ASCII-encoded texts to UTF-8 texts. With the help of this converter, we have converted approx. 45,000 words from different texts. Presently this corpus contains approx. one lakh words, with 10,196 sentences.

### IV. WORD ANALYSIS

We have analysed the word structure of the Bishnupriya Manipuri language from the data of the corpus. Some portions of our result are shown below:

1) NOUN:

**Gender Suffixes:**

A. In the case of words indicating human beings, the word মুনি (muni) and জেলা (ɉela) are used before the word to indicate masculine and feminine genders respectively.

E.g. –মুনিমানু (munimanu : *man*), জেলামানু (ʤelamanu : *woman*)

B. The feminine gender is generally indicated by the use of the word জেলা (ʤela) after the words indicating common gender.

C. Feminine gender is formed by adding the following suffixes to the masculine forms of words:

i) ী (i): খুড়া (kʰuɹa : *father's younger brother*) -> খুড়ী (kʰuɹi : *the wife of father's younger brother*), জেঠাবা (ʤetʰaba : *father's elder brother*) -> জেঠীমা (ʤetʰima : *the wife of father's elder brother*)

ii) নী (ani): চাকর (sakɔɹ : *servant*) -> চাকরানী (sakɔɹani : *maid servant*)

iii) নী (ni): চামার (samaɹ : *cobbler*) -> চামারনী (samaɹani : *female cobbler*)

**Number suffixes:**

A. The word গাছি (gaʃi) compounded with the stem preceding, which is invariably a noun of relationship, bears a plural sense.

E.g. – দাদাগাছি (dadagaʃi : *elder brothers*), মামাগাছি (mamagaʃi : *maternal uncles*)

B. গি (i) : This suffix is added to the singular forms in –গ (g) and হান (han) resulting to –গি (gi) and –হানি (hani) respectively, in the sense of '*a small number of*'.

E.g.- মাছগি (maʃɔgi : *the few fishes*)

C. মাহেই (mahei) : This suffix is added in the sense of 'a large number of'.

E.g.- গুরুমাহেই (guɹumahei : *many cows*)

D. In some cases, the plural sense is carried by the addition of adjectives bearing plural sense:

i) গুলি (guli): It is used as an adjective in the sense of 'a small number of'.

E.g. – মাছগুলি (maʃɔguli : *a few fishes*)

ii) হাবি (habi): It is used before or after the stem as an adjective in the sense of 'all'.

E.g. – মানুহাবি (manuhabi : *all men*), হাবিমানু (habimanu : *all men*)

iii) এতা (eta), ঔতা (outa): These are used after the stem as adjectives in the sense of 'these' and 'those' respectively.

E.g. – মানু এতা (manueta : *these people*)

**Case suffixes:**

A. Nominative: The nominative related to an intransitive verb generally does not take any ending. The terminations –এ (ɛ), -য (j) and –রে (ɯe), -যে (je) are generally added to the nominatives.

E.g. – পূর্ণয় ভাত খেইল (puɹnɔi bʰat kʰeil- : *Purna took his meals*)

B. Accusative: Objects indicating inanimate things generally do not take any termination. Objects indicating animals take the terminations –রে (ɹe) and –অরে (ɔɹe)

E.g. – মি মাণ্টুরে চাউরি (mi mantuɹe sauɹi : *I am looking at Mantu*)

C. Instrumental: The instrumental termination –ল (l) is added to the accusative form of the noun. As in the case of words indicating non-living things and lower animals, no termination is added in the accusative, the affix –ল (l) is directly added to the stem. In the case of words indicating higher animals, -ল (l) is added to the accusative form in –রে (ɹe) or –অরে (ɔɹe).

E.g. – মোরেল (mʊɹelɔ : *with me*)

D. Dative: -রে (ɹe), -অরে (ɔɹe), -রাঙ (ɹaŋ)

E.g. – মোরে দে (mʊɹe de : *give me*)

E. Ablative: -ত (tɔ), এত (etɔ) -তত (ttɔ), -এতত (ettɔ), -রাংত (ɹaŋtɔ)

E.g. – মোরাংত (mʊɹaŋtɔ : *from me*)

F. Genitive: The terminations are –র (ɹ) and –অর (ɔɹ). The affix – র (ɹ) becomes – অর (ɔɹ) when the stem ends in a consonant or has its final vowel unpronounced.

E.g. – মোর (mʊɹ : *my*)

G. Locative: -ত (t), -এ (e), -রাঙ (ɹaŋ)

E.g. - ঘরে (gʰɔɹe : *in the house*)

H. Vocative: -ও (ʊ), -রো (ɹʊ), -রে (ɹe)

E.g. – দাদারো (dadaɹʊ : *O elder brother*)

2) PRONOUN:
The Pronoun for the First Person:

A. Singular Form: মি (mi)

B. Plural Form: আমি (ami : *we*)

C. There are two oblique forms, মো (mʊ) and আমা (ama) to which the inflexions and post-positions are added to form various cases. মো (mʊ) is for the singular forms and আমা (ama) is for the plural forms.

E.g. – মো (mʊ) -> মোরে (mʊɹe), মোরেল (mʊɹel), মোরাং (mʊɹaŋ)

The Pronoun for the Second Person:

A. Singular Form: তি (ti)

B. Plural Form: তুমি (tumi)

C. There are oblique forms তো (tʊ) and –তুমা (tuma) to which case inflexions and post-positions are added to form various cases.

E.g. – তো (tʊ)-> তোরে (tʊɹe), তোরাং (tʊɹaŋ)

The Pronoun for the Third Person:

A. Singular Form: তা (ta) in masculine gender and তেই (tei) in feminine gender.

B. Plural Form: তানো (tanʊ) or তানু (tanu) which is common to both masculine and feminine genders.

C. The Proximate Demonstrative Pronoun: -এ (ɛ)

D. The Remote Demonstrative Pronoun: ঔ (ɔu)

E. The Relative Pronoun: জে (dʑe)

3) VERB:

There are three moods of verb in Bishnupriya: Indicative, Imperative and Subjunctive.

A. **The Indicative Mood**:

Indicative Mood may be divided into Simple and Compound.

The Simple Tense Suffixes:

1. Simple Present: -র (r), -উরি (uri), -ঔরি (ɔuɹi), -উরী (ɔuɹi), -ঔরী (ɔuɹi)

E.g. – কর (kɔɹ : *to do*) -> করর (kɔɹr), কররি (kɔɹuri)

2. The Precative Present: -ইতু (itu), -ইতে (ite), ইত (it), -ইতাং (itaŋ), -ইতাই (itai), -ইতা (ita)

The initial –i of the endings is dropped after roots ending in consonant. Initial –i- of the endings plus the final –i- of the root becomes –i-.

E.g. – কর (kɔɹ : *to do*) -> করতু (kɔɹtu), করতে (kɔɹte), করত (kɔt)

3. Simple Past: -ইলু (ilu), -ইলে (ile), -ইল (ilɔ), -ইলাং (ilaŋ), -ইলাই (ilaj), -ইলা (ila)

E.g. – পা (pa : *to get*) -> পেইলু (peilu), পেইলে (peile), পেইল (peilɔ)

4. Simple Future: -ইতৌ (itou), -ইতেই (itei), -ইতাউ (itau), -ইতাঙাই (itaŋai), -ইতারাই (itarai), -ইতাই (itai)

E.g. – খা (kʰa : *to eat*) -> খেইতৌ (kʰaitou), খেইতেই (kʰaitei), খেইতাউ (kʰaitau)

5. Probable Future: -ইম (im), -ইবে (ibe), -ইব (ibɔ), -ইবাং (ibaŋ), -ইবাই (ibaj), ইবা (iba)

E.g. – থ (tʰɔ : *to keep*) -> থইম (tʰɔim), থইবে (tʰɔibe), থইব (tʰɔibɔ)

The Compound Tense Suffixes:

1. Present Progressive: There is no special form for the present progressive tense. Its meaning is carried by the combination of the non-finite form of the principal verb in –ইয়া (ija) or –আত (at) / -নাত (nat) and the simple present tense of the root –আছ (aʃ).

E.g. - কর (kɔɹ : *to do*) -> করিয়া আছু (kɔɹija aʃu)

2. Present Perfect: -এছু (eʃu), -এছত (eʃɔt), -এছে (eʃe), -এছি (eʃi), -এছ (eʃ)

The initial –এ (e) of the endings plus the final –অ(ɔ) and –আ (a) of the root becomes -অ(ɔ) and -আ (a) respectively.

E.g. - থ (tʰɔ : *to keep*) -> থছু (tʰɔʃu)

3. Past Progressive: Its meaning is carried by the combination of the non-finite form of the principal verb in –ইয়া (ija) or –আত (at) / -নাত (nat) and the simple past tense form of the root –আছ (aʃ).

E.g. - পি (pi : *to drink*) -> পিয়া আছিলু (piya aʃilu)

4. Past Perfect: -এছিলু (eʃilu), -এছিলে (eʃile), -ইছিল (iʃil), -ইছিলাং (iʃilaŋ)

E.g. - পা (pa : *to get*) -> পাছিলে (paʃile)

5. Probable Past: It is formed by the combination of the non-finite form of the principal verb in –ইয়া (ija) or its contracted form –e and the probable future form the root –থা (tʰa : *to remain*).

E.g. - কর (kɔɹ : *to do*) -> করে থাইম (kɔɹe tʰaim)

6. Future Progressive: Its meaning is carried by the combination of the non-finite form of the principal verb in –ইয়া (ija) or –অত (ɔt) / -নাত (nat) and the simple future form of the root –থা (tʰa).

E.g. - কর (kɔɹ : *to do*) -> করিয়া থাইতৌ (kɔɹija tʰaitou )

B. **The Imperative Mood**:

The present imperative mood has the following endings: -ইং (iŋ), -অ (ɔ), -অক (ɔk), -ইক (ik), -অকা (ɔka)

E.g. - কর (kɔɹ : *to do*) -> করিং (kɔɹiŋ)

C. **The Subjunctive Mood**:

There is no special form for the subjunctive mood. Its meaning is carried by the help of words, such as, aiste and jadi, the first for the past subjunctive and the second for the future subjunctive.

V. **FREQUENCY ANALYSIS**

From the corpus, we have collected approx. twenty five thousands inflected words.

20 highest frequency words of the corpus are shown below:

TABLE I.  First twenty frequent words from the corpus

| Word | IPA : Meaning | Frequency |
|---|---|---|
| বারো | baɹo : *and, again* | 652 |
| আহান | ahan : *one* | 644 |
| বার | baɹ : *things* | 641 |
| আমার | amaɹ : *our* | 528 |
| নেই | nei : *absent* | 405 |
| তার | taɹ : *his* | 400 |
| না | na : *negative prefix* | 392 |
| লগে | lɔge : *with* | 369 |
| অয়া | ɔja : *becoming* | 366 |
| বুলিয়া | bulija : *having said* | 333 |
| করিয়া | kɔɹija : *having done* | 303 |
| করে | kɔɹe : *having done* | 302 |
| দিয়া | dija : *giving* | 297 |
| এহান | ehan : *this* | 290 |
| আর | aɹ : *other* | 266 |
| তেই | tei : *she* | 262 |
| হাবি | habi : *all* | 256 |
| অকরল | ɔkɹɔlɔ : *started* | 254 |
| মানু | manu : *man* | 254 |
| আছে | aʃe : *remain* | 250 |

## VI.  CONCLUSION AND FUTURE WORKS

In this paper, we have discussed the development of a text-based corpus in the Bishnupriya Manipuri language. Due to the lack of awareness like other Indian languages, this language is studied less frequently. As a result, the language still lacks a good corpus and basic language processing tools. We are unaware of any corpus development work on Bishnupriya Manipuri language. Our main motivation is to develop the resources of linguistics work on this language.

In future, we will further increase the size of this corpus and will add more sections to the corpus. We are also planning to develop language processing tools on this language. One interesting morphological feature of this language is the word formation with Tibeto-Burman Meitei roots. Studies of such special features remain to be done in future.